\def\year{2022}\relax
\documentclass[letterpaper]{article} 
\usepackage{aaai22}  
\usepackage{times}  
\usepackage{helvet}  
\usepackage{courier}  
\usepackage[hyphens]{url}  
\usepackage{graphicx} 
\usepackage{natbib}  
\usepackage{caption} 
\DeclareCaptionStyle{ruled}{labelfont=normalfont,labelsep=colon,strut=off} 
\usepackage{algorithm}
\usepackage{algorithmic}

\usepackage{newfloat}
\usepackage{listings}
\floatstyle{ruled}
\newfloat{listing}{tb}{lst}{}
\floatname{listing}{Listing}
\usepackage{amsfonts} 
\usepackage{amssymb} 
\usepackage{pifont}
\usepackage{arydshln} 
\usepackage{xcolor} 
\usepackage{mathtools} 
\usepackage{adjustbox} 
\usepackage{multirow} 
\usepackage{paralist} 
\usepackage{CJKutf8} 
\usepackage{booktabs} 

\usepackage{enumerate}
\usepackage{bbding} 
\usepackage{wasysym}

\usepackage{subcaption}

\usepackage{amsthm} 
\theoremstyle{definition} 

\usepackage{tikz}
\usepackage[edges]{forest} 

\usepackage{custom} 

\usepackage{hyperref}

\title{
  \textit{Text is no more Enough}!
  \\ A Benchmark for Profile-based Spoken Language Understanding
}
\author{
    Xiao Xu\textsuperscript{\rm 1}\thanks{\ Equal Contribution.},
    Libo Qin\textsuperscript{\rm 1}\footnotemark[1], 
    Kaiji Chen\textsuperscript{\rm 2}, 
    Guoxing Wu\textsuperscript{\rm 2}, 
    Linlin Li\textsuperscript{\rm 2}, 
    Wanxiang Che\textsuperscript{\rm 1}\thanks{\ Email Corresponding.}
}
\affiliations {
    \textsuperscript{\rm 1}Research Center for Social Computing and Information Retrieval \\
    \textsuperscript{\rm 1}Harbin Institute of Technology, China \\
    \textsuperscript{\rm 2}Huawei Technologies Co., Ltd. \\ 
    \{xxu, lbqin, car\}@ir.hit.edu.cn,
    \{chenkaiji, wuguoxing1, lynn.lilinlin\}@huawei.com
}

\begin{document}
\maketitle
\begin{abstract} \label{Abstract}
	Current researches on spoken language understanding (SLU) heavily are limited to a simple setting: the plain text-based SLU that takes the user utterance as input and generates its corresponding semantic frames (e.g., intent and slots). 
	Unfortunately, such a simple setting may fail to work in complex real-world scenarios when an utterance is semantically ambiguous, which cannot be achieved by the text-based SLU models.
    In this paper, we first introduce a new and important task, \textbf{Pro}file-based \textbf{S}poken \textbf{L}anguage \textbf{U}nderstanding (\textsc{ProSLU}), which requires the model that not only relies on the plain text but also the supporting profile information to predict the correct intents and slots.
    To this end, we further introduce a large-scale human-annotated Chinese dataset with over 5K utterances and their corresponding supporting profile information (\textit{Knowledge Graph} (KG), \textit{User Profile} (UP), \textit{Context Awareness} (CA)).
    In addition,  we evaluate several state-of-the-art baseline models and explore a multi-level knowledge adapter to effectively incorporate profile information. 
    Experimental results reveal that all existing text-based SLU models fail to work when the utterances are semantically ambiguous and our proposed framework can effectively fuse the supporting information for \textit{sentence-level} intent detection and \textit{token-level} slot filling.
    Finally, we summarize key challenges and provide new points for future directions, which hopes to facilitate the research.
\end{abstract}
\section{Introduction}
\label{Introduction}
Spoken Language Understanding (SLU)~\citep{young2013pomdp,ijcai2021-622} is a core component in task-oriented dialogue systems, aiming to extract intent and semantic constituents from the natural language utterances~\citep{tur2011spoken}.
It consists of two typical subtasks: intent detection and slot filling to map the user input utterance into an overall intent and a slot label sequence.

\begin{figure}[t]
	\centering
	\includegraphics[width=0.47\textwidth]{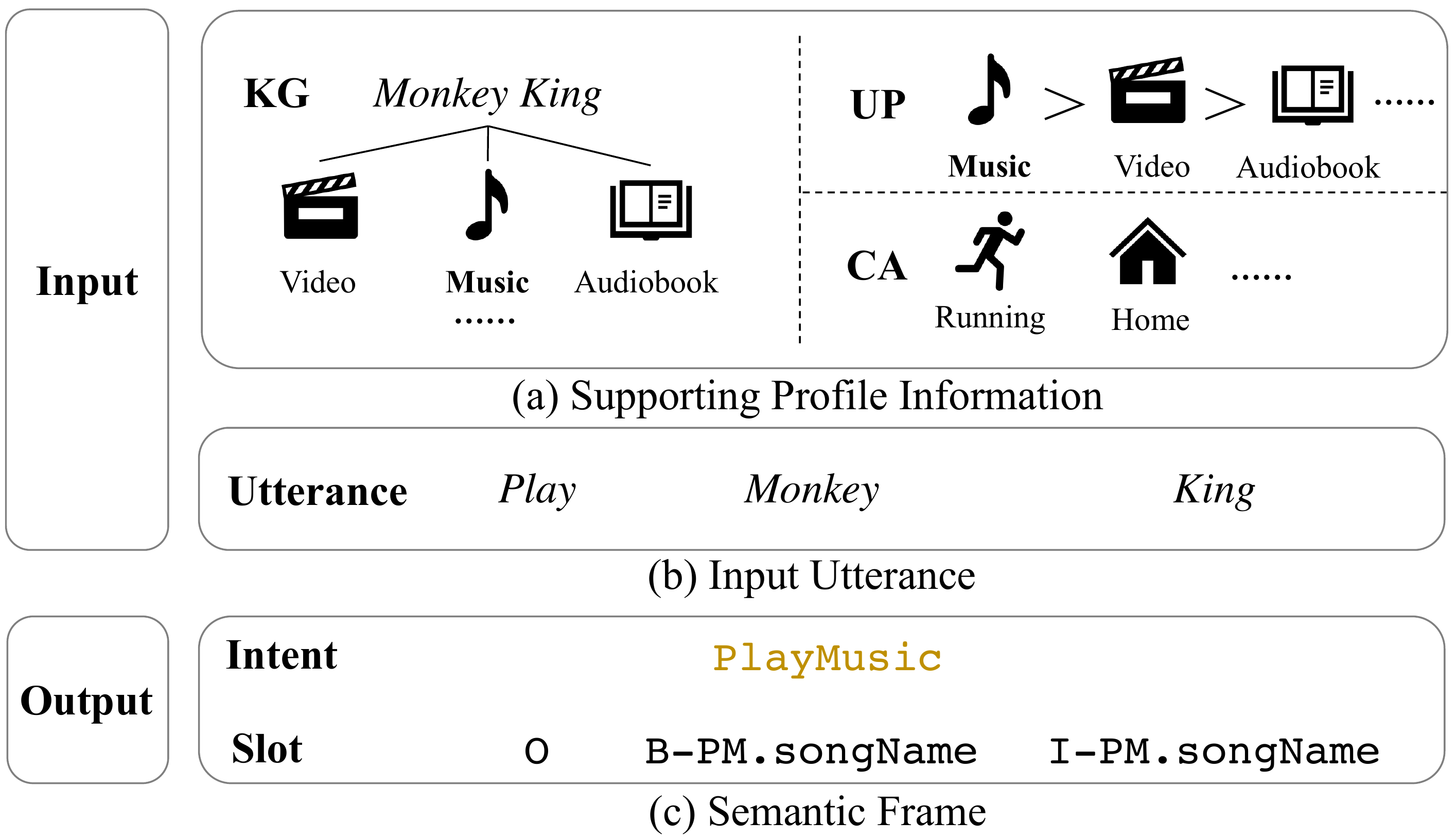}
	\caption{An example in \textsc{ProSLU}. Semantic frame denotes the intent and slots of the input utterance. The ``$>$''  implies that the probability of the former is greater than the latter.
	}
	\label{fig:example-simple}
\end{figure}

With the help of pre-trained models, recent work has achieved remarkable success on the SLU system.
~\citet{ijcai2021-622} surveys that performance improvement on traditional SLU is relatively already saturated because the neural joint model has achieved over 96\% and 99\% on slot filling and intent detection on the ATIS dataset~\citep{hemphill-etal-1990-atis}.
Though achieving good performance, the current researches on SLU mainly focus on a simple scenario: the plain text-based setting.

More specifically, traditional SLU systems are based on the assumption that simply relying on plain text can capture intent and slots correctly.
Unfortunately, such an assumption may not be achieved in real-world scenarios when the user utterance is semantically ambiguous.
For example, as shown in \figurename~\ref{fig:example-simple}(b)\&(c), when a user asks the agent (e.g., Apple Siri) for the query ``\textit{Play {Monkey King}}'', simply relying on the text is not enough for extracting correct semantic frame results, since ``\textit{Monkey King}'' could indicate a ``\texttt{rock song}'' or an ``\texttt{eponymous Chinese TV cartoon}''. 
\textit{Therefore, we argue that the existing text-based SLU is not enough for the complex setting in real-world scenarios when the utterance is semantically ambiguous.}
In this paper, we assume that the profile information of the user can help to solve the issue where the corresponding profile information can be used as supplementary knowledge to alleviate the ambiguity of utterance.
As illustrated in \figurename~\ref{fig:example-simple}(a), if a user is running and prefers music to video, \textit{Monkey King} is more likely to be music than video.
Unfortunately, none of the work considers the profile-based SLU in real-world scenarios.
One of the key reasons for hindering the progress is the lacking of public benchmarks.

In the paper, to bridge the research gap, we propose a new and important task, \textbf{Pro}file-based \textbf{S}poken \textbf{L}anguage \textbf{U}nderstanding (\textsc{ProSLU}), which requires a model not only depends on the text but also on the given supporting profile information.
We further introduce a Chinese human-annotated dataset, with over 5K utterances annotated with intent and slots, and corresponding supporting profile information. 
In total, we provide three types of supporting profile information: 
(1) \textit{Knowledge Graph} (KG) consists of entities with rich attributes, 
(2) \textit{User Profile} (UP) is composed of user settings and information,
(3) \textit{Context Awareness} (CA) is user state and environmental information.

To establish baselines on \textsc{ProSLU}, we evaluate several state-of-the-art models. The experimental results reveal that all models fail to work (lower than 50\% in overall accuracy metric) on \textsc{ProSLU}.
In addition, we propose a \textit{multi-level} knowledge adapter to equip the existing SLU models with the ability to incorporate profile information, which has the following advantages: 
(1) it achieves a fine-grained knowledge injection for both \textit{sentence-level} intent detection and \textit{token-level} slot filling; 
(2) it can be used as a plugin and easily be compatible with the existing state-of-the-art SLU models.

Contributions of this work are concluded as: 
\begin{itemize}
	\item We systematically analyze the state-of-the-art SLU models and observe that existing models fail to work in real-world scenarios, which shed a light for future research.
	\item 
	We propose a new and important task named \textsc{ProSLU}. In addition, we introduce a Chinese human-annotated dataset, hoping it would push forward further research. 
	To our knowledge, we are the first to explore \textsc{ProSLU}.
	\item We establish various baselines and conduct qualitative analysis for \textsc{ProSLU}. 
	Besides, we explore a \textit{multi-level} adapter to effectively inject the profile information.
\end{itemize}

We hope the task and datasets will invite more research on \textsc{ProSLU}. All datasets and codes used in this paper are publicly available at \url{https://github.com/LooperXX/ProSLU}.
\begin{table}[t]
	\centering
	\begin{adjustbox}{width=0.48\textwidth}
		\begin{tabular}{l|l}
			\hline 
			\multicolumn{2}{c}{\textbf{Input}} \\ \hline
			\textbf{Utterance} & \textit{Play Monkey King} \\ \hline
			\textbf{KG} & \texttt{Mention} ``\textit{Monkey King}'': \{music, video and audiobook\}, ...\\ \hline
			\textbf{UP} & \texttt{Preference} for [music, video, audiobook]: {[}0.5, 0.3, 0.2{]},  ... \\ \hline
			\textbf{CA} & \texttt{Movement State}: Running, \ \texttt{Geographic Location}: Home, ... \\ \hline
			\hline
			\multicolumn{2}{c}{\textbf{Output}} \\ \hline
			\textbf{Intent}    & \texttt{PlayMusic} \\ \hline
			\textbf{Slot}      & \texttt{O B-PlayMusic.songName I-PlayMusic.songName} \\ \hline
		\end{tabular}
	\end{adjustbox}
	\caption{
		A simplified example from the \textsc{ProSLU} dataset.
	}
	\label{tab:example-simple}
\end{table}

\section{Problem Definition}
\label{PD}
In this section, we define the supporting profile information, profile-based intent detection and slot filling.
A simplified example is given in \tablename~\ref{tab:example-simple} and the complete example can be found in the Appendix~\ref{appendix:example}.

\subsection{Supporting Profile Information}
\label{sec:info_definition}
We introduce three types of supporting profile information including \textit{Knowledge Graph}, \textit{User Profile} and \textit{Context Awareness}, which are used to help the model to alleviate the ambiguity in the utterances.

\begin{figure*}[t]
	\centering
	\includegraphics[width=0.87\textwidth]{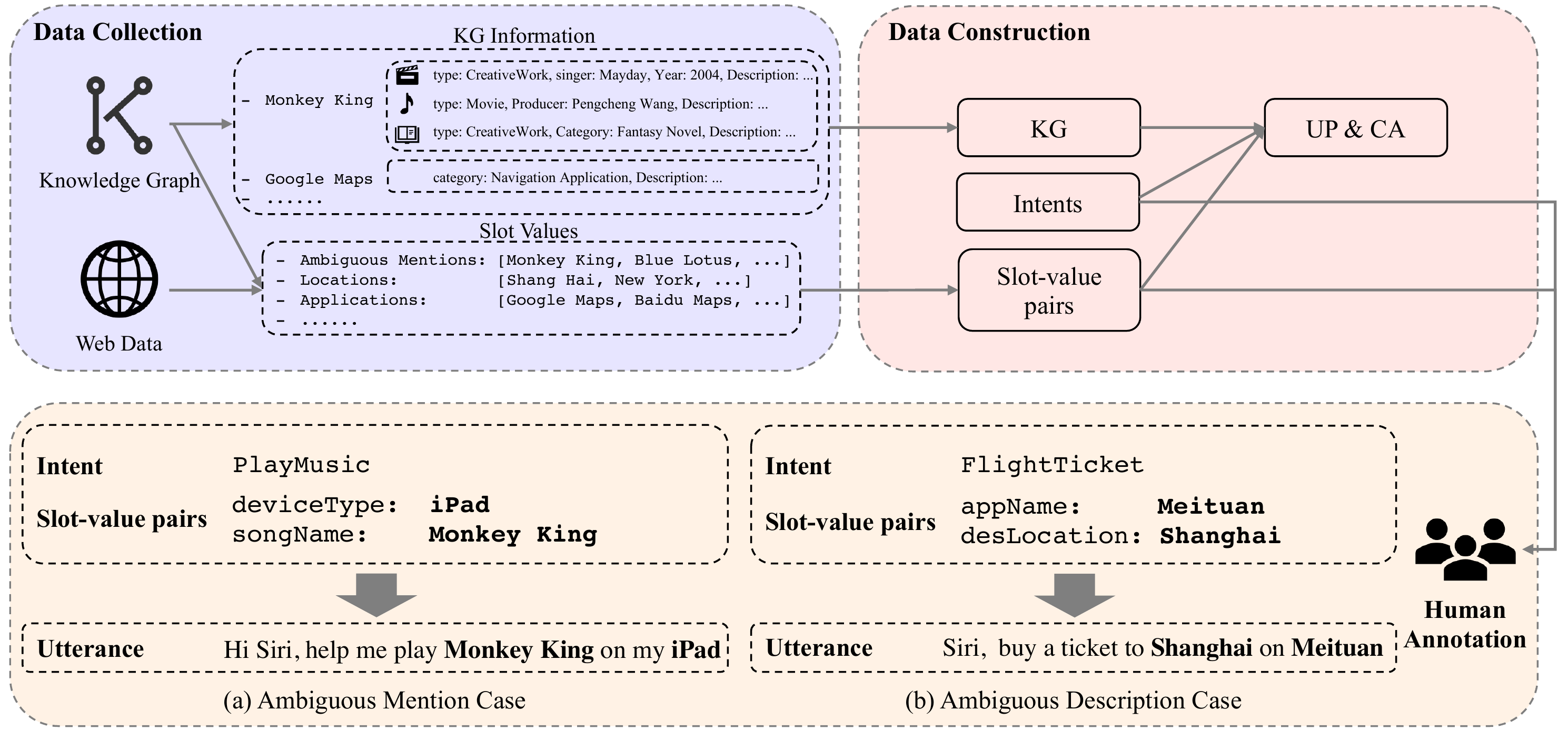}
	\caption{
        The overall workflow of our data collection, data construction and human annotation. 
    }
	\label{fig:workflow}
\end{figure*}

\paragraph{Knowledge Graph}
The first type of profile information is the \textbf{K}nowledge \textbf{G}raph (KG), which contains large amounts of interlinked entities and their corresponding rich attributes.\footnote{We use open-source encyclopedia knowledge graphs like {CN-DBpedia}, {OwnThink}, etc.}
Depending on the context, an ambiguous mention refers to some different entities of the same (or similar) name but different entity types, as the ambiguous mention (or their shared name) tends to be polysemous (i.e., have multiple meanings).
Take \tablename~\ref{tab:example-simple} for example, KG information provides background knowledge for the ambiguous mention \textit{Monkey King} (e.g., it can be an entity of music, video or audiobook).
Following~\citet{Chen2020TabFact}, we represent each entity and its attributes in KG as a long text sequence, which is composed of key-value pairs (e.g., ``\texttt{subject}: \textit{Monkey King}, \texttt{type}: CreativeWork''). 

\paragraph{User Profile}
The second type of profile information is the \textbf{U}ser \textbf{P}rofile (UP), 
which is a collection of settings and information (items) associated with the user.
Each item in UP consists of a non-negative float array that sums to 1.
As shown in \tablename~\ref{tab:example-simple}, the user ``preferences for music, video, and audiobook: $[0.5, 0.3, 0.2]$'' is an item in UP information, which can help the model to judge that the user prefers listening to music rather than watching videos. 
We concat all the items in UP and directly flatten them to a single feature vector $\mathbf{x}_{\text{UP}} \in \mathbb{R}^{u}$ ($u$ is the UP feature dimension).
For example, the user preferences for music, video, and audiobook are $[0.5, 0.3, 0.2]$ and the user transportation preferences for subway, bus and driving are $[0.4, 0.1, 0.5]$, we could get $[0.5, 0.3, 0.2, 0.4, 0.1, 0.5]$.

\paragraph{Context Awareness}
The third type of information is the \textbf{C}ontext \textbf{A}wareness (CA) that denotes the user state and environmental information,
including the user's movement state, posture, geographic location, etc.
As shown in \tablename~\ref{tab:example-simple}, a user who is running is more likely to play music than video. 
The form of each item in CA is similar to UP, e.g., the movement state can be walking, running, or stationary, and [0,1,0] indicates that the movement state is running
Similarly, we get the flatten feature vector $\mathbf{x}_{\text{CA}} \in \mathbb{R}^{c}$ ($c$ is the CA feature dimension).

\subsection{Profile-based Intent Detection and Slot Filling}
\label{sec:definition-SLU}
Unlike the traditional SLU task, \textsc{ProSLU} requires the model to predict results not only rely on the input utterance, but also the corresponding supporting profile information.

Specifically, given an input word sequence $\mathbf{x} = (x_1, \dots, x_T)$ ($T$ is the number of words) and its corresponding supporting profile information, 
profile-based intent detection can be seen as a sentence classification problem to decide the intent label $o^{\mathrm{I}}$ while profile-based slot filling is a sequence labeling task to generate a slot sequence $\mathbf{o}^{\mathrm{S}} = (o_1^{\mathrm{S}}, \dots, o_T^{\mathrm{S}})$.
Formally, the \textsc{ProSLU} task can be defined as:
\begin{equation}
	(o^{\mathrm{I}}, \mathbf{o}^{\mathrm{S}}) = f(\mathbf{x}, \text{KG}, \text{UP}, \text{CA}),
\end{equation}
where $f$ denotes the trained model.

\section{Dataset}
\label{Dataset}
In this section, we describe the collection and annotation process of the \textsc{ProSLU} dataset. 
In \textsc{ProSLU}, each utterance is semantically ambiguous, which requires the model to leverage the supporting profile information.
\subsection{Ambiguity Definition}
\label{sec:definition-Ambiguity}
By manually collecting semantically ambiguous samples from the error cases in real-world systems, we found that
there are two main sources of ambiguity in user utterances in real-world scenarios, including ambiguous mentions and ambiguous descriptions.
\paragraph{Ambiguous Mentions}
It indicates that the presence of ambiguous mentions in the user utterance introduces lexical ambiguity and ultimately leads to semantic ambiguity in the utterance.
For example, the ambiguous mention \textit{Monkey King} can represent different entities such as a rock song (sung by the Mayday band), a biographical novel, or an eponymous Chinese TV cartoon.

\paragraph{Ambiguous Descriptions} 
It indicates that the ambiguity is caused by the ambiguous semantic understanding of utterance rather than the ambiguous mentions.
Take the user utterance ``I want to buy a ticket to Shanghai'' as an example, it's hard to capture the correct intent simply depending on the utterance, because the intent of utterance could be to book a train ticket, a plane ticket, or a coach ticket.

\subsection{Data Design}
\paragraph{Intent and Slot Design}
We design multiple ambiguous intent groups based on real-world scenarios and open-source SLU datasets.
For example, \{\texttt{PlayMusic}, \texttt{PlayVideo}, \texttt{PlayAudioBook}\} is an ambiguous intent group where each intent is ambiguous with each other (``Play Monkey King'' can be any intent in this ambiguous intent group).
Slot labels are collected directly from the slot label sets corresponding to the intents.
\paragraph{Supporting Profile Information Design}
For KG information, we collect it directly from the open-source knowledge graphs.
For UP and CA information, based on the UP and CA schemas in real scenarios,
we carefully pick out the items that can help disambiguate the above intents and slot labels, and integrate them to form the final UP and CA items. 

\subsection{Data Collection}
We first collect values for different slots based on the crawled public web data.
Then we collect ambiguous mentions from the knowledge graph.
The entities could share the same mentions but have different entity types in the knowledge graph.
For example, the song entity \textit{Monkey King} and the Chinese TV cartoon entity \textit{Monkey King} has the same mention (name) but different entity types. 
Therefore, we are able to collect numerous ambiguous mentions.

\subsection{Data Construction}
Based on the first two steps, we design the data generation process separately for these two ambiguity cases defined 
in Section~\ref{sec:definition-Ambiguity}.
\subsubsection{Ambiguous Description Case}
For each sample, we first randomly select an intent and some slots in its corresponding slot label sets.
Next, we fill these slots by randomly selecting slot values from the collected slot values.
In addition, when some slot values are entities in the knowledge graph, we extract KG information for them. 
Finally, we design the heuristic rules to generate valid UP and CA information for the corresponding intent.
For example, when the selected intent is \texttt{SearchDriveRoute} (search for driving routes between two places), 
the ``\texttt{Has Car}'' item in UP information is more likely to be ``true'' and 
the ``\texttt{Movement State}'' in CA information may be less likely to be ``on the aircraft''.\footnote{Users who do have a car are usually more likely to ask for driving routes, and users who do not have an Internet connection on the aircraft are usually less likely to try to search.}

\subsubsection{Ambiguous Mention Case}
To bring lexical ambiguity into the utterance, there should exist ambiguous mentions in the slot values of the utterance.
Thus, slightly different from the former case, after randomly selecting the intent and slots, 
the ambiguous mention should be randomly selected but satisfies the selected intent.\footnote{For example, the selected mention must have a music entity to satisfy the \texttt{PlayMusic} intent.}
Then we generate slot-value pairs and obtain the KG information of different entities corresponding to the selected mention from the knowledge graph data.
Based on the selected intent and the entities in the KG information, we design the hard-coded heuristics to randomly generate valid UP and CA information.
For example, for the \texttt{PlayVideo} intent, the movement state in CA information is less likely to be ``running'', and if the entity types of entities in the KG information are \{music, video and audiobook\},
the user preferences for video tend to greater than music and audiobook.

\begin{table}[t]
	\centering
	\begin{adjustbox}{width=0.3\textwidth}
		\begin{tabular}{l|r}
			\hline \hline
			\#Utterances & 5,249 \\
			\#Utterances in Train Set & 4,196 \\ 
			\#Utterances in Valid Set & 522 \\ 
			\#Utterances in Test Set & 531 \\ 
			\#Avg. Words per Utterance & 23.64 \\ 
			\#Intents & 14\\
			\#Slots & 99\\
			\#UP & 4\\
			\#CA & 4\\
			\#KG entities & 7,466\\
			\#Avg. KG Entity & 2.77 \\
			\#Avg. Words per Entity & 272.63\\ 
			\hline \hline
		\end{tabular}
	\end{adjustbox}
	\caption{
		Data statistics of \textsc{ProSLU} dataset. \#Avg. KG Entity denotes the average number of entity per data sample.
	}
	\label{tab:statistics}
\end{table}

\subsection{Human Annotation}

After data collection and construction, 
the annotators only need to manually write the ambiguous utterances in conjunction with the given intent and slot-value pairs. 

We hire an annotation team to check the generated data in each given sample and to annotate the utterances.
More importantly, the utterances annotated by the annotators must be reasonable and logical but semantically ambiguous.
The sample with unreasonable generated data will be removed. 
\figurename~\ref{fig:workflow} gives an illustration of the overall workflow.

\subsection{Quality Control}
To ensure quality, each sample is annotated by three experts and the annotation process lasts for nearly two months. 
In practice, we randomly divide all the completed annotated samples into 10 groups and select 50 sentences from each group for testing, and if more than 5 sentences are regarded as incorrectly annotated, the whole group would be re-annotated. 
Finally, we obtain 5,249 samples, where the ratio of description ambiguity vs. mention ambiguity case in the dataset is nearly 1:2. 
\tablename~\ref{tab:statistics} summarizes the detailed statistics of the \textsc{ProSLU} dataset.
\section{Approach}
\label{Approach}
In this section, we first introduce the general SLU model and then describe the proposed multi-level knowledge adapter, which can be used for \textit{sentence-level} intent detection and \textit{word-level} slot filling, respectively.\footnote{The detailed description and training objectives of the general SLU model can be found in the Appendix~\ref{appendix:general-slu-model}.}

\subsection{General SLU Model} \label{sec:general_model}
The general SLU model consists of a shared encoder, an intent detection decoder, and a slot filling decoder.
\subsubsection{Shared Encoder}
The shared encoder reads the input utterance $\mathbf{x}=\{{x}_{1}, {x}_{2},.., {x}_{T}\}$ ($T$ is the number of tokens in the input utterance) to generate the shared encoding representation 
$\mathbf{E}=\{\mathbf{e}_{1}, \mathbf{e}_2, \dots, \mathbf{e}_{T}\}\!=\!\operatorname{Encoder}\left({x}_{1}, {x}_{2},.., {x}_{T}\right).$

\begin{figure}[t]
	\centering
	\includegraphics[width=0.47\textwidth]{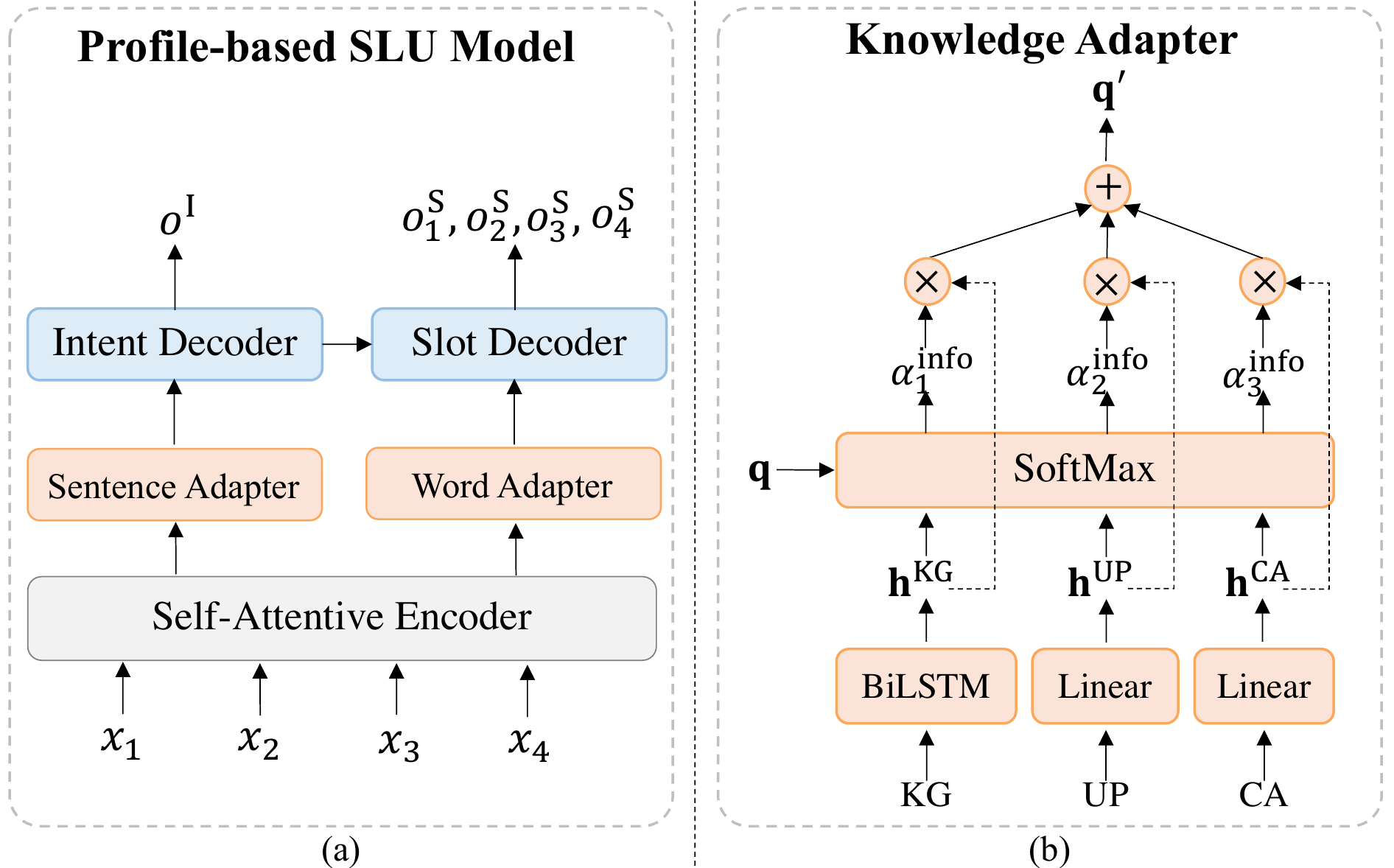}
	\caption{
		The illustration of Profile-based SLU model (a), which consists of a general SLU model in Section~\ref{sec:general_model} and a knowledge adapter (b) in Section~\ref{sec:knowledge_adapter}.
    }
	\label{fig:framework}
\end{figure}

\subsubsection{Intent Detection Decoder}
Based on the shared encoding representation $\mathbf{E}$, 
a sentence representation $\mathbf{g}$ can be generated (e.g., the sentence self-attention mechanism~\citep{zhong-etal-2018-global}) for intent detection:
\begin{eqnarray}
    {\mathbf{y}}^{\mathrm{I}} &=& \operatorname{softmax}\left({\mathbf{W}}_{\mathrm{I}} \, \mathbf{g} \right), \label{eq:intent} \\
    {o}^{\mathrm{I}} &=& \arg \max ({\mathbf{y}}^{\mathrm{I}}),
\end{eqnarray}
where $\mathbf{W}_{\mathrm{I}}$ are trainable parameters. 

\subsubsection{Slot Filling Decoder}
The unidirectional LSTM is used as the slot filling decoder. 
At each decoding step $t$, we adopt the intent-guided mechanism~\citep{qin-etal-2019-stack} and the decoder state $\mathbf{h}^{\mathrm{S}}_{t}$ is a function of 
the previous decoder state ${\mathbf{h}}_{t-1}^{\mathrm{S}}$,
the aligned encoder hidden state $\mathbf{e}_t$, 
the embedding of predicted intent and previously emitted slot.

Finally, $\mathbf{h}^{\mathrm{S}}_{t}$ is used for slot filling:
\begin{eqnarray}
    {\mathbf{y}}_{t}^{\mathrm{S}} &=& \operatorname{softmax} \left({\mathbf{W}}_{\mathrm{S}} \, {\mathbf{h}}_{t}^{\mathrm{S}}\right),\label{eq:slot} \\
    {o}_{t}^{\mathrm{S}} &=& \arg \max ({\mathbf{y}}_{t}^{\mathrm{S}}),
\end{eqnarray}
where ${o}_{t}^{\mathrm{S}}$ is the slot label of the $t$-th word in the utterance and $\mathbf{W}_{\mathrm{S}}$ are trainable parameters.

\subsection{Supporting Information Representations} 
\paragraph{KG Representation}
KG information of each entity is a concatenated sequence of key-value pairs. 
For KG information with only one entity, 
e.g., $\mathbf{x}^{\text{KG}} = \{x^{\text{KG}}_1, x^{\text{KG}}_2, \dots, x^{\text{KG}}_L\}$ ($L$ is the number of words in the KG sequence),
we use BiLSTM to obtain the KG encoding representations $\mathbf{H}^{\text{KG}}=\{\mathbf{h}_{1}^{\text{KG}}, \mathbf{h}_2^{\text{KG}}, \dots, \mathbf{h}_{L}^{\text{KG}}\} \in \mathbb{R}^{L \times d_i}$ by applying $\mathbf{h}_t^{\text{KG}} = \operatorname{BiLSTM}\left(\phi^{\text{KG}}\left(x^{\text{KG}}_{t}\right) , \mathbf{h}_{t-1}^{\text{KG}}\right)$, where $d_i$ is the information embedding dimension.
We directly use the last hidden state $\mathbf{h}_L^{\text{KG}}$ as the KG representation $\mathbf{h}^{\text{KG}}$.
For KG information consisting of sequences of multiple entities, we perform average pooling for the last hidden state of each sequence as the overall aggregated KG representation $\mathbf{h}^{\text{KG}}$.

\paragraph{UP and CA Representation}
UP and CA representations can be obtained by using linear projection, using $\mathbf{h}^{\text{UP}} = \mathbf{W}_\text{UP}^\top \, \, \mathbf{x}_{\text{UP}}$ and $\mathbf{h}^{\text{CA}} = \mathbf{W}_\text{CA}^\top \, \mathbf{x}_{\text{CA}}$, respectively, where $\mathbf{W}_\text{UP} \in \mathbb{R}^{u \times d_i}$ and $\mathbf{W}_\text{CA} \in \mathbb{R}^{c \times d_i}$ are trainable parameters.

\subsection{Multi-level Knowledge Adapter}
\label{sec:knowledge_adapter}
\subsubsection{Knowledge Adapter}
The core challenge of \textsc{ProSLU} is how to effectively incorporate the supporting information.
We explore a knowledge adapter to address this challenge, which can be used as a plugin without changing the original SLU structure. 
Inspired by~\citet{Srinivasan2020MultimodalSR}, we adopt the hierarchical attention fusion mechanism~\citep{luong-etal-2015-effective, libovicky-helcl-2017-attention} as the knowledge adapter, which has the advantage of dynamically considering relevant supporting information for different words.
Specifically, given the query vector $\mathbf{q}$ and the corresponding supporting information $\mathbf{H}^{\text{Info}} = [\mathbf{h}^{\text{KG}};\mathbf{h}^{\text{UP}};\mathbf{h}^{\text{CA}}]\in \mathbb{R}^{3 \times d_i}$, we obtain the updated representation $ \mathbf{q}'$ = $\text{Knowledge-Adapter} (\mathbf{q}, \mathbf{H}^{\text{info}}) $ by weighted summing the representation from all the supporting information:
\begin{eqnarray}
    \alpha_{i}^{\text{info}} = \frac{\operatorname{exp}\left( \mathbf{q} \, \mathbf{W} \, \mathbf{h}^{\text{info}}_i \right)}{\sum_{k=1}^3 \operatorname{exp}\left( \mathbf{q} \, \mathbf{W} \, \mathbf{h}^{\text{info}}_k \right)}, \\
    	 \mathbf{q}' = \sum_{i=1}^3 \alpha_{i}^{\text{info}} \mathbf{h}^{\text{info}}_i,
    	  \label{eq:hierarchical1} 
 \label{eq:hierarchical2}
\end{eqnarray}
where $\mathbf{W}$ are trainable parameters and $\{\mathbf{h}_1^{\text{Info}}, \mathbf{h}_2^{\text{Info}}, \mathbf{h}_3^{\text{Info}}\}$ denotes $\{\mathbf{h}^{\text{KG}},\mathbf{h}^{\text{UP}},\mathbf{h}^{\text{CA}}\}$ respectively.
\subsubsection{Sentence-level Knowledge Adapter for Intent Detection}
\begin{table*}[ht]
	\centering
	\begin{adjustbox}{width=0.95\textwidth}
		\begin{tabular}{l|c|c|c|c|c|c}
			\hline \hline
			\multirow{2}{*}{\textbf{Model}} & \multicolumn{3}{c|}{\textbf{w/o Profile}}& \multicolumn{3}{c}{\textbf{w/ Profile}}\\ \cline{2-7}
			& {Slot (F1)}  & {Intent (Acc)}   & {Overall (Acc)} & {Slot (F1)}  & {Intent (Acc)}   & {Overall (Acc)} \\ 
			\hline 
			\multicolumn{7}{c}{Non Pre-trained SLU Models} \\
			\hline
			Slot-Gated~\citep{goo-etal-2018-slot} & 36.53 & 41.24 & 32.02 & 74.18 & 83.24 & 69.11 \\
			Bi-Model~\citep{wang-etal-2018-bi}& 37.37 & 44.63 & 32.58 & 77.76 & 82.30 & 73.45 \\
			SF-ID~\citep{e-etal-2019-novel} & 39.63 & 42.37 & 30.89 & 73.70 & 83.24 & 68.36 \\
			Stack-Propagation~\citep{qin-etal-2019-stack} & 39.29 & 39.74 & 36.35 & 81.08 & 83.99 & 78.91 \\
			General SLU Model & 42.24 & 43.13 & 37.85 & 83.27 & 85.31 & 79.10 \\
			\hline 
            \multicolumn{7}{c}{Pre-trained-based SLU Models} \\
            \hline
			BERT~\citep{devlin-etal-2019-bert} & 44.80 & 45.76 & 42.18 & 82.51 & 84.56 & 80.98 \\
			XLNet~\citep{yang2019xlnet} & \textbf{46.92} & \textbf{48.59} & \textbf{43.88} & 83.39 & 85.88 & 81.73 \\
			RoBERTa~\citep{liu2019roberta} & 45.92 & 47.83 & 43.13 & 82.90 & 85.31 & 81.17 \\
			ELECTRA~\citep{clark2020electra} & 46.48 & 47.46 & 42.56 & \textbf{84.38} & \textbf{86.63} & \textbf{82.30} \\
            \hline \hline
		\end{tabular}
	\end{adjustbox}
	\caption{
		Slot Filling and Intent Detection results on the \textsc{ProSLU} dataset.
	}
	\label{tab:results}
\end{table*}

We perform a sentence-level knowledge adapter for sentence-level intent detection, where we use sentence representation $\mathbf{g}$ as query to obtain the hierarchical fused information $\mathbf{s}^\text{info}$, using $\mathbf{s}^\text{info}$ = $\operatorname{Knowledge-Adapter} (\mathbf{g}, \mathbf{H}^{\text{info}})$, which is used for augmenting intent detection:
\begin{equation}
    {\mathbf{y}}^{\mathrm{I}} = \operatorname{softmax}\left({\mathbf{W}}_{\mathrm{I}}\left( \mathbf{g} \oplus \mathbf{s}^\text{info} \right) \right).
\end{equation}
\subsubsection{Word-level Knowledge Adapter for Slot Filling}
Since slot filling is a word-level sequence labeling task, we apply a word-level knowledge adapter to inject different relevant knowledge for each word.

Specifically, we use the self-attentive encoding $\mathbf{e}_{t}$ at the $t$-th timestep as query vector to fuse supporting information using $\mathbf{s}_t^\text{info}$ =  $\operatorname{Knowledge-Adapter} (\mathbf{e}_{t}, \mathbf{H}^{\text{info}})$.
Similarly, $\mathbf{s}_t^\text{info}$ is used to enhance word-level representation in slot filling decoder:
\begin{eqnarray}
    {\mathbf{h}}_{t}^{\mathrm{S}} &=& \operatorname{LSTM} \left(\mathbf{s}_t \oplus \mathbf{s}_t^\text{info},\mathbf{h}^{\mathrm{S}}_{t-1} \right)\\
	{\mathbf{y}}_{t}^{\mathrm{S}} &=& \operatorname{softmax} \left({\mathbf{W}}_{\mathrm{S}} {\mathbf{h}}_{t}^{\mathrm{S}}\right).
\end{eqnarray}
where $\mathbf{s}_t$ is the concatenation of the aligned encoder hidden state, intent embedding, and the previous slot embedding. 
\section{Experiments}
\label{experiments}
\subsection{Experimental Settings}
The self-attentive encoder hidden units are $256$ in all datasets. $\ell_2$ regularization is $1 \times 10^{-6}$ and the dropout rate is $0.4$ for reducing overfitting. We use Adam~\citep{kingma2014adam} to optimize the parameters in our model and adopt the suggested hyper-parameters for optimization. For all the experiments, we select the model which works best on the dev set and then evaluate it on the test set. All experiments are performed on the GPU Tesla V100. 

\subsection{Baselines}
We experiment the existing state-of-the-art non pre-trained SLU models on the \textsc{ProSLU} dataset: 
1) {\texttt{Slot-Gated Atten.}}~\citet{goo-etal-2018-slot} proposes a slot-gated joint model to explicitly model the correlation between slot filling and intent detection. 
2) {\texttt{Bi-Model.}}~\citet{wang-etal-2018-bi} proposes the Bi-model to study the cross-impact between the intent detection and slot filling. 
3) {\texttt{SF-ID Network.}}~\citet{e-etal-2019-novel} proposes an SF-ID network to construct direct connections for the slot filling and intent detection. 
4) {\texttt{Stack-Propagation}}~\citet{qin-etal-2019-stack} adopts a joint model with Stack-Propagation to capture the intent semantic knowledge.
We also investigate the existing state-of-the-art multi-intent models on \textsc{ProSLU} dataset in the Appendix~\ref{sec:multi-intent}.

To investigate the impact of pre-trained models in our \textsc{ProSLU} dataset, based on the general SLU model.
we adopt the pre-trained models \texttt{BERT}~\citep{devlin-etal-2019-bert}, \texttt{XLNet}~\citep{yang2019xlnet}, \texttt{RoBERTa}~\citep{liu2019roberta}, \texttt{ELECTRA}~\citep{clark2020electra} as the shared encoder to get the pre-trained-based SLU models.

\subsection{Analysis on Baselines without Profile Information}
Following~\citet{goo-etal-2018-slot} and ~\citet{qin-etal-2019-stack}, we evaluate the performance of slot filling using F1 score, intent detection using accuracy, the sentence-level semantic frame parsing using overall accuracy which represents all metrics are right in an utterance.
\paragraph{Non Pre-Trained SLU Models Performance}
We conduct experiments on \textsc{ProSLU} to observe the performance of the non pre-trained SLU models without supporting profile information.
The results are shown in \tablename~\ref{tab:results}. 
We observe that all baseline models significantly drop a lot compared with ATIS and SNIPS dataset on all three metrics.
For example, the baseline model Stack-Propagation~\citep{qin-etal-2019-stack} achieved 86.5\% and 86.9\% on overall accuracy on ATIS and SNIPS but only obtain 36.35\% on the \textsc{ProSLU} dataset.\footnote{Note that, compared with General SLU Model, Stack-Propagation use token-level intent detection instead of sentence-level intent detection, which brings performance degradation. We speculate that because the input unit of all experiments is the character, token-level intent detection may bring some noise.}
This indicates that the existing models fail to work when the utterances are semantically ambiguous. 

\paragraph{Pre-Trained-based SLU Models Performance}
In this section, we further conduct experiments with the pre-trained-based SLU models on \textsc{ProSLU} to observe its performance without supporting profile information.
The same trend is observed in \tablename~\ref{tab:results}. 
The overall accuracy of all pre-trained-based SLU models (w/o Profile) is still less than 45\%, which indicates that simply using pre-trained models does not alleviate the situation.

\paragraph{Comparison Between Non Pre-Trained Models and Pre-Trained Models}
Comparing with non pre-trained SLU models, we can see that all the pre-trained-based SLU models are better than the non pre-trained SLU models, which can bring 4\% to 6\% improvement.
We attribute this to the fact that the pre-trained models learn general semantic knowledge in the pre-training stage, hence it can provide rich semantic features that can help to ease the ambiguity in \textsc{ProSLU}.

\subsection{Analysis on Baselines with Profile Information}
\tablename~\ref{tab:results} (w/ Profile column) shows the performance of all models with supporting profile information on the \textsc{ProSLU} dataset.
It can be seen that the performance of all models improve significantly by a large margin based on our multi-level knowledge adapter to incorporate supporting profile information.
All the three metrics improve by about 30\% to 40\%, which indicates the supporting profile information can help alleviate the ambiguity in the ambiguous utterances.
It further proves the significance and importance of our \textsc{ProSLU} task for real-world scenarios.

\subsection{Ablation Study of Multi-level Knowledge Adapter}
To explore the effectiveness of multi-level knowledge adapter for \textsc{ProSLU} task, we perform the ablation study on the best model, ELECTRA-based SLU model (w/ Profile), in \tablename~\ref{tab:results-ablation}.

\begin{table}[t]
	\centering
	\begin{adjustbox}{width=0.47\textwidth}
		\begin{tabular}{lccc}
			\toprule
            \textbf{Model}
			& \textbf{Slot (F1)}  & \textbf{Intent (Acc)}   & \textbf{Overall (Acc)} \\ 
			\midrule
            ELECTRA & \textbf{84.38} & \textbf{86.63} & \textbf{82.30} \\ 
			\hdashline
            ~~w/o Sentence-level Adapter & 77.32 & 48.78 & 43.88  \\
			~~w/o Word-level Adapter & 79.99 & 81.36 & 78.15 \\
			~~w/o Multi-level Adapter & 46.48 & 47.46 & 42.56 \\ 
            \bottomrule
		\end{tabular}
	\end{adjustbox}
	\caption{Ablation Study of Multi-level Knowledge Adapter.}
	\label{tab:results-ablation}
\end{table}

\paragraph{Effect of Sentence-level Adapter}
We first experiment by only adopting a word-level slot adapter and the results are shown in \tablename~\ref{tab:results-ablation} (w/o Sentence-level Adapter Row), we observe a significant decrease in all three metrics, 7.06\%, 37.85\%, 38.42\%, respectively, and performance degradation is most obvious on intent detection task.
This is because the sentence-level intent adapter can effectively help the intent detection task to identify the correct intent and transfer the correct guidance knowledge for the slot filling task through explicitly interacting between two tasks. 

\paragraph{Effect of Word-level Adapter}
We remove the word-level slot adapter and only adopt the sentence-level intent adapter, which means there is no direct supporting information is injected into the slot filling decoder.
The results are shown in \tablename~\ref{tab:results-ablation} (w/o Word-level Adapter Row).
We observe that our framework drops 4.39\% in slot filling task, which indicates that the word-level adapter can effectively inject profile knowledge for word-level slot filling task.
An interesting observation is that the performance decrease of three metrics is slightly lower compared to w/o Sentence-level Adapter. We assume that although we remove the world-level adapter, the remaining sentence-level adapter can still help train a good intent detector, which can be used to guide the slot filling task with the intent-guided mechanism we adopted.

\paragraph{Effect of Multi-level Adapter}
We remove the proposed multi-level adapter and directly conduct experiments with the ELECTRA-based SLU model (w/o Profile).
As shown in \tablename~\ref{tab:results-ablation} (w/o Multi-level Adapter Row), we observe a more significant decrease in all three metrics, 37.90\%, 39.17\%, 39.74\%, respectively.
This further demonstrates that our multi-level adapter can effectively incorporate supporting profile information into the intent detection task and slot filling task, achieving fine-grained knowledge transfer to effectively cope with ambiguous sentences in real scenarios.

\subsection{Error Analysis}
In this section, we empirically provide error samples of two different types generated from ELECTRA-based SLU model(w/ Profile).

\subsubsection{KG Representation}
When KG information consists of seven entities, it is relatively large and become hard to be represented and understood by the model. 
As shown in the first block of \tablename~\ref{tab:error-analysis}, when the KG information shows the ``\textit{Answer}'' can be the music or audiobook entity, the model predicts ``\textit{Answer}'' as ``\texttt{O}'' incorrectly.
We attribute it to the fact that we simply represent each entity by flatting its KG information into a sequence and perform average pooling to obtain the overall aggregated KG representation, which would not effectively and correctly represent the KG information.

\subsubsection{Supporting Profile Information Fusion}
Take the second block in \tablename~\ref{tab:error-analysis} as an example, although the supporting profile information shows that the user prefers listening to \textit{audiobook} rather than \textit{watching videos}, the model predicts the intent as \texttt{PlayVideo} incorrectly, which means the knowledge fusion between the three types of supporting profile information needs to be more accurate and effective.

\begin{table}[t]
	\centering
	\begin{adjustbox}{width=0.48\textwidth}
		\begin{tabular}{l|l}
			\hline \hline 
			\textbf{Utterance} & \textit{Open my iPad and search for Answer , sung by Joey Yung} \\ 
			\hdashline
			\textbf{Supporting} & \textbf{KG}: \texttt{Mention} ``\textit{Answer}'': \textbf{7} entities of\{music or audiobook\} \\
			\hdashline
			\multirow{2}{*}{\textbf{Slot}} 
			& Predict:	\texttt{O O deviceType O O O songName O artist artist} \\
			& Real:	\texttt{\quad O O deviceType O O O \quad \ \  O \quad \ \  O artist artist} \\
			\hline 
			\hline 
			\textbf{Utterance} & \textit{Play Martial Universe by Hu Li on my iPad} \\ 
			\hdashline
			\multirow{3}{*}{\textbf{Supporting}} 
			& \textbf{KG}: Mention ``\textit{Martial Universe}'': \textbf{3} entities of \{video or audiobook\}, ...\\ 
			& \textbf{UP}: \texttt{Preference} for [music, video, audiobook]: {[}0.4, 0.2, 0.4{]},  ... \\
			& \textbf{CA}: \texttt{Movement State}: Walking, \texttt{Geographic Location}: Home, ... \\ 			
			\hdashline
			\multirow{2}{*}{\textbf{Intent}} 
			& Predict: \texttt{PlayVideo} \\
			& Real: \texttt{\quad PlayAudioBook} \\
			\hline \hline 
		\end{tabular}
	\end{adjustbox}
	\caption{
		Error examples in ELECTRA-based SLU model (w/ Profile). 
		Some information in KG, UP and CA are omitted for brevity.
	}
	\label{tab:error-analysis}
\end{table}

\subsection{Challenges}
Based on above analysis, we summarize the current challenges for our \textsc{ProSLU} dataset.
\paragraph{Representation of KG Information}
In this paper, we follow~\citet{Chen2020TabFact} to represent KG information as long text sequences which composed of key-value pairs. 
Numerous entities with plentiful attributes poses a huge challenge for the KG encoder. 
It is an important and fundamental issue to investigate more efficient ways to encode KG information.
For example, it is also possible to train the representation of each entity directly on the knowledge graph through knowledge graph embedding methods~\citep{NIPS2013_1cecc7a7}.

\paragraph{Effectiveness of Fusion Approaches}
We follow~\citet{libovicky-helcl-2017-attention} to adopt the hierarchical attention fusion mechanism to fuse three types of supporting profile information.
Many representation fusion approaches exist in the machine learning research area~\citep{zadeh-etal-2017-tensor, liu-etal-2018-efficient, tsai-etal-2019-multimodal}.
It will be challenging and rewarding to explore the effectiveness of these approaches on \textsc{ProSLU}.

\paragraph{Expansion of Supporting Profile Information}
In this paper, we investigate three types of supporting profile information, which is common in real-world scenarios. 
To better solve \textsc{ProSLU} task and alleviate ambiguity from user utterances, more types of supporting profile information can be expanded in future research.

\section{Related Work}
\label{related work}
Dominant SLU systems adopt the joint model to jointly consider the correlation between intent detection and slot filling~\cite{ijcai2021-622}. 
~\citet{zhang2016joint} and ~\citet{hakkani2016multi} propose the multi-task framework to jointly model the correlation between the two tasks.
\citet{goo-etal-2018-slot,qin-etal-2019-stack} and~\citet{Teng2021InjectingWI} explicitly incorporate intent information for guiding the slot filling.
Another series of work ~\citep{li-etal-2018-self,e-etal-2019-novel, qin2021co} consider the bidirectional connection between the two tasks.
\citet{zhu2020dual} also considers the semi-supervised NLU setting.
However, their work mainly focus on the plain text-based SLU. In contrast, we mainly consider the ambiguous setting and propose a new and important task, \textsc{ProSLU} which requires a model to predict the intent and slots correctly given text and its supporting profile information. 

The ambiguous problem has attracted increasing attention in dialogue direction. 
~\citet{bhargava2013easy,xu2014contextual,chen2015leveraging,chen2016end,su-etal-2018-time,qin2021knowing} have shown leveraging contextual information can handle the ambiguous problem in SLU direction.
Compared with their work, we focus on how to incorporate the corresponding supporting profile information to alleviate ambiguity in a single-turn setting while they adopt the multi-turn interaction manner.
Another strand of work~\citet{zhang-etal-2018-personalizing, zheng2019personalized, song-etal-2020-profile} consider incorporating profile information to ease ambiguity and generate consistent dialogue responses.
Unlike their work, we focus on the SLU domain while they mainly consider the end-to-end dialogue systems.
To the best of our knowledge, this is the first work to consider additional information to alleviate the ambiguity of utterances in the SLU system.
\section{Conclusion}
\label{conclusion}
In this paper, we investigate the Profile-based SLU, which requires a model to rely not only on the surface utterance but also on the supporting information.
We further introduce a large-scale annotated dataset to facilitate further research.
In addition, we explore a multi-level knowledge adapter to effectively inject the supporting information.
To the best of our knowledge, we are the first to consider Profile-based SLU.

\appendix

\section*{Ethical Considerations}
Each sample in our dataset is checked by annotators to ensure the content does not pose potential risks. As indicated in the main text, the annotators are properly paid, and we’ve taken several steps to both ensure a proper working burden and a high quality dataset.

\begin{table}[t]
    \centering
    \begin{adjustbox}{width=0.47\textwidth}
        \begin{tabular}{l}
        \toprule
        \textbf{Intent Group} \\ 
        \midrule
        PlayMusic, PlayVideo, PlayAudioBook \\
        SearchMusic, SearchVideo, SearchAudioBook \\
        SearchLocation, SearchLocationOntheway \\
        SearchMetroRoute, SearchBusRoute, SearchDriveRoute \\
        SearchTrainTicket, SearchFlightTicket, SearchCoachTicket \\ 
        \bottomrule
        \end{tabular}
    \end{adjustbox}
    \caption{
        Intent groups in the \textsc{ProSLU} dataset.
    }
    \label{tab:intent-group}
\end{table}

\begin{figure*}[t]
    \centering
    \includegraphics[width=0.9\textwidth]{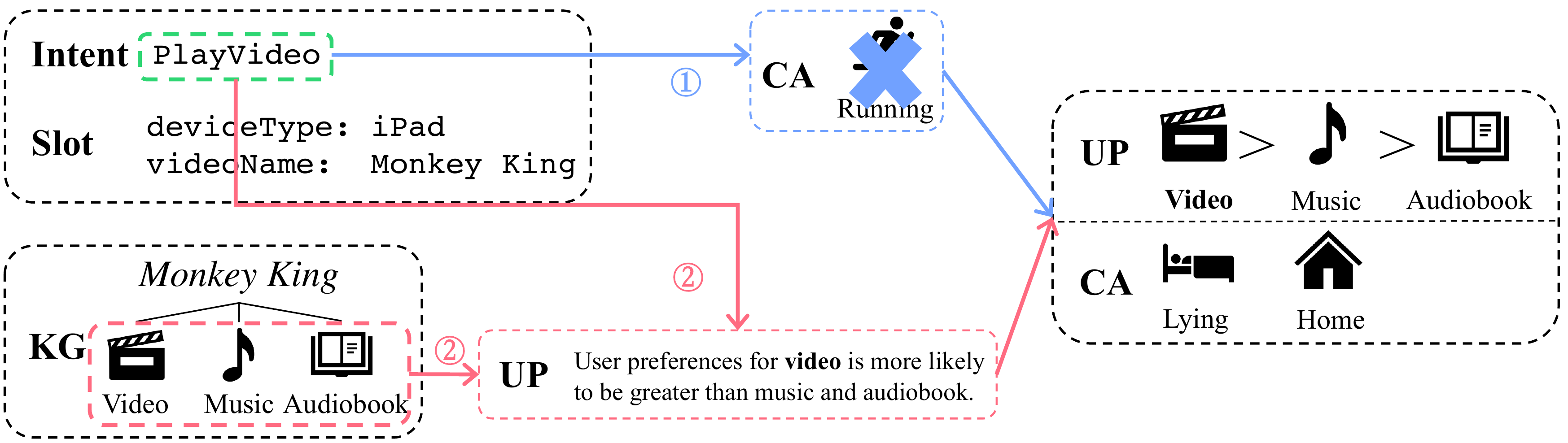}
    \caption{
        An illustration of generating UP and CA Information.
    }
    \label{fig:generate}
\end{figure*}

\begin{table*}[t]
    \centering
    \begin{adjustbox}{width=0.9\textwidth}
        \begin{tabular}{l|l}
        \hline \hline 
        \multicolumn{2}{c}{\textbf{Input}} \\ \hline
        \textbf{Utterance} & \textit{Play Monkey King} \\ \hline
        \multirow{3}{*}{\textbf{KG}} & 
        \texttt{subject}: \textit{Monkey King}, \ \texttt{type}: CreativeWork, \ \texttt{singer}: Mayday, \ \texttt{Year}: 2004, \texttt{Description}: ..., \\ 
        & \texttt{subject}: \textit{Monkey King}, \ \texttt{type}: Movie, \ \texttt{Producer}: Pengcheng Wang, \texttt{Writer}: Pengcheng Wang, \ \texttt{Description}: ...,  \\ 
        & \texttt{subject}: \textit{Monkey King}, \ \texttt{type}: CreativeWork, \ \texttt{Category}: Fantasy Novel, \ \texttt{Description}: ..., \\ \hline
        \textbf{UP} & \texttt{Preference} for [music, video \& audiobook]: {[}0.2, 0.7, 0.1{]}, \texttt{Has Car}: True, ... \\ \hline
        \textbf{CA} & \texttt{Movement State}: stationary, \ \texttt{Posture}: lying down, \ \texttt{Geographic Location}: home, ... \\ \hline
        \multicolumn{2}{c}{\textbf{Output}} \\ \hline
        \textbf{Intent}    & \texttt{PlayVideo} \\ \hline
        \textbf{Slot}      & \texttt{O B-PlayVideo.videoName I-PlayVideo.videoName} \\ \hline
        \hline
        \end{tabular}
    \end{adjustbox}
    \caption{
        An example in the \textsc{ProSLU} dataset.
    }
    \label{tab:example}
\end{table*}

\section{Appendix}

\subsection{Example}
\label{appendix:example}
\subsubsection{Ambiguous Intent Groups}
The ambiguous intent groups we designed are shown in \tablename~\ref{tab:intent-group}. 
For example, \{\texttt{PlayMusic}, \texttt{PlayVideo}, \texttt{PlayAudioBook}\} is an ambiguous intent group where each intent is ambiguous with each other (``Play Monkey King'' can be any intent in this ambiguous intent group).
Slot labels are collected directly from the slot label sets corresponding to the intents.

\subsubsection{An Illustration of Generating UP and CA Information}
As shown in \figurename~\ref{fig:generate}, for the \texttt{PlayVideo} intent, the movement state in CA information cannot be running, and if the entity types of entities in the KG information are music, video and audiobook,
the user preferences for video is more likely to be greater than music and audiobook.

\subsubsection{A Detailed Example in \textsc{ProSLU}}
We show a detailed example in our \textsc{ProSLU} dataset in \tablename~\ref{tab:example}. 
For the utterance ``Play Monkey King'', \texttt{PlayMusic, PlayVideo, PlayAudioBook} are possible intents in our intent set. 
Given the three entities in the KG information, the above intents are all reasonable. 
Considering the UP information, we can find that the user likes watching videos more than listening to music and audiobook. Finally, the CA information shows that the user is lying at home, which is a reasonable state to watch videos. Therefore, the real intent of the user can be predicted to \texttt{PlayVideo}.

\subsection{General SLU model}
\label{appendix:general-slu-model}
\subsubsection{Intent Detection Decoder}
To perform intent detection, a sentence self-attention mechanism~\citep{zhong-etal-2018-global} is applied for obtaining sentence representation $\mathbf{g}$, using:
\begin{eqnarray}
    \alpha_{i} &=& \frac{\operatorname{exp}\left( \mathbf{w}_\mathrm{g}^\top \mathbf{e}_i \right)}{\sum_{j} \operatorname{exp}\left( \mathbf{w}_\mathrm{g}^\top \mathbf{e}_j \right)}, \\
    \mathbf{g} &=& \sum\nolimits_{i} \alpha_{i} \mathbf{e}_i,
\end{eqnarray}
where $\mathbf{w}_g \in \mathbb{R}^{d} $ are trainable model parameters.

Then, $\mathbf{g}$ is used as input for intent detection:
\begin{eqnarray}
    {\mathbf{y}}^{\mathrm{I}} &=& \operatorname{softmax}\left({\mathbf{W}}_{\mathrm{I}} \ \mathbf{g} \right), \label{eq:intent} \\
    {o}^{\mathrm{I}} &=& \arg \max ({\mathbf{y}}^{\mathrm{I}}),
\end{eqnarray}
where ${o}^{\mathrm{I}}$ is the predicted intent label; $\mathbf{W}_{\mathrm{I}}$ are trainable parameters. 

\subsubsection{Slot Filling Decoder}
We use a unidirectional LSTM as the slot filling decoder.
At each decoding step $t$, we adopt the intent-guided mechanism~\citep{qin-etal-2019-stack} and the decoder hidden state $ \mathbf{h}^{\mathrm{S}}_t $  can be formalized as:
\begin{eqnarray}
    \mathbf{s}_t &=& \mathbf{e}_t \oplus \phi^{\text{intent}} \left({o}^{\mathrm{I}} \right) \oplus \phi^{\text{slot}}\left({o}^{\mathrm{S}}_{t-1}\right), \\
    {\mathbf{h}}_{t}^{\mathrm{S}} &=& \operatorname{LSTM} \left(\mathbf{s}_t,\mathbf{h}^{\mathrm{S}}_{t-1} \right),
\end{eqnarray}
where ${\mathbf{h}}_{t-1}^{\mathrm{S}}$ is the previous decoder state;
$\mathbf{e}_t$ is the aligned encoder hidden state
and $\mathbf{s}_t$ is the concatenated input for the slot filling decoder; $\phi^{\text{intent}}\left(\cdot\right)$ and $\phi^{\text{slot}}\left(\cdot\right)$ represent the embedding matrix of intents and slots, respectively. 
Finally, $\mathbf{h}^{\mathrm{S}}_{t}$ is used for slot filling:
\begin{eqnarray}
    {\mathbf{y}}_{t}^{\mathrm{S}} &=& \operatorname{softmax} \left({\mathbf{W}}_{\mathrm{S}} {\mathbf{h}}_{t}^{\mathrm{S}}\right),\label{eq:slot} \\
    {o}_{t}^{\mathrm{S}} &=& \arg \max ({\mathbf{y}}_{t}^{\mathrm{S}}),
\end{eqnarray}
where ${o}_{t}^{\mathrm{S}}$ is the slot label of the $t$-th word in the utterance and $\mathbf{W}_{\mathrm{S}}$ are trainable parameters.

\subsubsection{Joint Training}
The intent detection objection is formulated as:
\begin{equation}
  \mathcal{L}_\mathrm{I} \triangleq-\sum_{i=1}^{n_\mathrm{I}}\hat{{{y}}}^{i,\mathrm{I}} \log \left({{y}}^{i,\mathrm{I}}\right).
\end{equation}

Similarly, the slot filling task objection is defined as: 
\begin{equation}
\mathcal{L}_\mathrm{S} \triangleq-\sum_{t=1}^{T}\sum_{i=1}^{n_\mathrm{S}}{\hat{{{y}}}_{t}^{i,\mathrm{S}}\log \left( {{y}}_{t}^{i,\mathrm{S}}\right)},
\end{equation}
where ${\hat{{{y}}}^{i,\mathrm{I}}}$ and ${\hat{{{y}}}_{t}^{i,\mathrm{S}}}$ are golden intent labels and golden slot labels separately, $n_\mathrm{I}$ and $n_\mathrm{S}$ is the number of intent labels and slot labels respectively.

The final joint objective to optimize intent detection and slot filling together is formulated as:
\begin{equation}
  \mathcal{L} = \mathcal{L}_\mathrm{I} + \mathcal{L}_\mathrm{S}.
\end{equation}

In addition, the shared encoding representations learned by the shared self-attentive encoder can consider two tasks jointly and further ease the error propagation compared with pipeline models~\citep{zhang2016joint} through the final joint loss function. 

\begin{table}[h]
	\centering
	\begin{adjustbox}{width=0.47\textwidth}
		\begin{tabular}{lccc}
			\toprule
            \textbf{Models}
			& \textbf{Slot (F1)}  & \textbf{Intent (Acc)}   & \textbf{Overall (Acc)} \\ 
            \midrule
			AGIF~\citep{qin-etal-2020-agif} & 42.55 & 36.35 & 33.90 \\
			AGIF w/Profile & 80.57 & 81.54 & 74.95 \\
            \midrule
            GL-GIN~\citep{qin-etal-2021-gl} & 37.49 & 38.61 & 35.03\\
			GL-GIN w/Profile & 82.70 & 85.69 & 79.28 \\
            \bottomrule
		\end{tabular}
	\end{adjustbox}
	\caption{
        The Performance of Multi-Intent Baselines on the \textsc{ProSLU} dataset.
    }
	\label{tab:results-multi}
\end{table}

\subsection{Exploration of the Multi-Intent Baselines} \label{sec:multi-intent}
We explore the state-of-the-art multi-intent baselines AGIF~\citep{qin-etal-2020-agif} and GL-GIN~\citep{qin-etal-2021-gl} on the \textsc{ProSLU} dataset.
The results are shown in \tablename~\ref{tab:results-multi}. 
We observe that both baselines show poor performance without profile information. With the help of supporting profile information, they improve significantly by a large margin based on our multi-level knowledge adapter.

\subsection{Ablation Experiments of Different Fusion Methods}

In addition to adopting the hierarchical attention fusion mechanism in Section~\ref{sec:knowledge_adapter}, we also try to utilize two traditional fusion layers to aggregate information from different sources as knowledge adapter:
\begin{itemize}
    \item Concatenation (Concat) is a simple and effective method~\citep{wu2018improving} that directly concatenate representation from different sources for each sample and
    \item Multilayer Perceptron (MLP) can automatically capture the integrated representation~\citep{nguyen2018improved} which applies an MLP layer on the concatenated output to further abstract the expressive aggregated representations and better extract the multi-source information.
\end{itemize}

As shown in \tablename~\ref{tab:results-ablation-fusion}, MLP fusion achieves better results than Concat fusion, 
but underperforms our hierarchical fusion method. This demonstrates that our hierarchical fusion can get word-level dynamic representations of multi-source information and inject them through the multi-level adapter to achieve fine-grained knowledge transfer.

\begin{table}[h]
	\centering
	\begin{adjustbox}{width=0.47\textwidth}
		\begin{tabular}{lccc}
			\toprule
            \textbf{Fusion Methods}
			& \textbf{Slot (F1)}  & \textbf{Intent (Acc)}   & \textbf{Overall (Acc)} \\ \midrule
			{Concat} & 80.08 & 82.67 & 76.84 \\
            {MLP} & 80.60 & 83.43 & 77.78 \\
            {Hierarchical} & \textbf{83.27} & \textbf{85.31} & \textbf{79.10} \\ 
            \bottomrule
		\end{tabular}
	\end{adjustbox}
	\caption{Ablation Experiments of Different Fusion Methods.}
	\label{tab:results-ablation-fusion}
\end{table}

\section*{Acknowledgements}
This work was supported by the National Key R\&D Program of China via grant 2020AAA0106501 and the National Natural Science Foundation of China (NSFC) via grant 61976072 and 61772153. This work was also supported by the Zhejiang Lab’s International Talent Fund for Young Professionals. 

\bibliography{custom}

\begin{thebibliography}{41}
\providecommand{\natexlab}[1]{#1}

\bibitem[{Bhargava et~al.(2013)Bhargava, Celikyilmaz, Hakkani-T{\"u}r, and
  Sarikaya}]{bhargava2013easy}
Bhargava, A.; Celikyilmaz, A.; Hakkani-T{\"u}r, D.; and Sarikaya, R. 2013.
\newblock Easy contextual intent prediction and slot detection.
\newblock In \emph{2013 IEEE International Conference on Acoustics, Speech and
  Signal Processing}, 8337--8341. IEEE.

\bibitem[{Bordes et~al.(2013)Bordes, Usunier, Garcia-Duran, Weston, and
  Yakhnenko}]{NIPS2013_1cecc7a7}
Bordes, A.; Usunier, N.; Garcia-Duran, A.; Weston, J.; and Yakhnenko, O. 2013.
\newblock Translating Embeddings for Modeling Multi-relational Data.
\newblock In Burges, C. J.~C.; Bottou, L.; Welling, M.; Ghahramani, Z.; and
  Weinberger, K.~Q., eds., \emph{Advances in Neural Information Processing
  Systems}, volume~26. Curran Associates, Inc.

\bibitem[{Chen et~al.(2020)Chen, Wang, Chen, Zhang, Wang, Li, Zhou, and
  Wang}]{Chen2020TabFact}
Chen, W.; Wang, H.; Chen, J.; Zhang, Y.; Wang, H.; Li, S.; Zhou, X.; and Wang,
  W.~Y. 2020.
\newblock TabFact: {A} Large-scale Dataset for Table-based Fact Verification.
\newblock In \emph{8th International Conference on Learning Representations,
  {ICLR} 2020, Addis Ababa, Ethiopia, April 26-30, 2020}. OpenReview.net.

\bibitem[{Chen et~al.(2016)Chen, Hakkani-T{\"u}r, T{\"u}r, Gao, and
  Deng}]{chen2016end}
Chen, Y.-N.; Hakkani-T{\"u}r, D.; T{\"u}r, G.; Gao, J.; and Deng, L. 2016.
\newblock End-to-end memory networks with knowledge carryover for multi-turn
  spoken language understanding.
\newblock In \emph{Interspeech}, 3245--3249.

\bibitem[{Chen et~al.(2015)Chen, Sun, Rudnicky, and
  Gershman}]{chen2015leveraging}
Chen, Y.-N.; Sun, M.; Rudnicky, A.~I.; and Gershman, A. 2015.
\newblock Leveraging behavioral patterns of mobile applications for
  personalized spoken language understanding.
\newblock In \emph{Proceedings of the 2015 ACM on International Conference on
  Multimodal Interaction}, 83--86.

\bibitem[{Clark et~al.(2020)Clark, Luong, Le, and Manning}]{clark2020electra}
Clark, K.; Luong, M.; Le, Q.~V.; and Manning, C.~D. 2020.
\newblock {ELECTRA:} Pre-training Text Encoders as Discriminators Rather Than
  Generators.
\newblock In \emph{8th International Conference on Learning Representations,
  {ICLR} 2020, Addis Ababa, Ethiopia, April 26-30, 2020}. OpenReview.net.

\bibitem[{Devlin et~al.(2019)Devlin, Chang, Lee, and
  Toutanova}]{devlin-etal-2019-bert}
Devlin, J.; Chang, M.-W.; Lee, K.; and Toutanova, K. 2019.
\newblock {BERT}: Pre-training of Deep Bidirectional Transformers for Language
  Understanding.
\newblock In \emph{Proceedings of the 2019 Conference of the North {A}merican
  Chapter of the Association for Computational Linguistics: Human Language
  Technologies, Volume 1 (Long and Short Papers)}, 4171--4186. Minneapolis,
  Minnesota: Association for Computational Linguistics.

\bibitem[{E et~al.(2019)E, Niu, Chen, and Song}]{e-etal-2019-novel}
E, H.; Niu, P.; Chen, Z.; and Song, M. 2019.
\newblock A Novel Bi-directional Interrelated Model for Joint Intent Detection
  and Slot Filling.
\newblock In \emph{Proceedings of the 57th Annual Meeting of the Association
  for Computational Linguistics}, 5467--5471. Florence, Italy: Association for
  Computational Linguistics.

\bibitem[{Goo et~al.(2018)Goo, Gao, Hsu, Huo, Chen, Hsu, and
  Chen}]{goo-etal-2018-slot}
Goo, C.-W.; Gao, G.; Hsu, Y.-K.; Huo, C.-L.; Chen, T.-C.; Hsu, K.-W.; and Chen,
  Y.-N. 2018.
\newblock Slot-Gated Modeling for Joint Slot Filling and Intent Prediction.
\newblock In \emph{Proceedings of the 2018 Conference of the North {A}merican
  Chapter of the Association for Computational Linguistics: Human Language
  Technologies, Volume 2 (Short Papers)}, 753--757. New Orleans, Louisiana:
  Association for Computational Linguistics.

\bibitem[{Hakkani-T{\"u}r et~al.(2016)Hakkani-T{\"u}r, T{\"u}r, Celikyilmaz,
  Chen, Gao, Deng, and Wang}]{hakkani2016multi}
Hakkani-T{\"u}r, D.; T{\"u}r, G.; Celikyilmaz, A.; Chen, Y.-N.; Gao, J.; Deng,
  L.; and Wang, Y.-Y. 2016.
\newblock Multi-domain joint semantic frame parsing using bi-directional
  rnn-lstm.
\newblock In \emph{Interspeech}, 715--719.

\bibitem[{Hemphill, Godfrey, and Doddington(1990)}]{hemphill-etal-1990-atis}
Hemphill, C.~T.; Godfrey, J.~J.; and Doddington, G.~R. 1990.
\newblock The {ATIS} Spoken Language Systems Pilot Corpus.
\newblock In \emph{Speech and Natural Language: Proceedings of a Workshop Held
  at Hidden Valley, {P}ennsylvania, June 24-27,1990}.

\bibitem[{Kingma and Ba(2014)}]{kingma2014adam}
Kingma, D.~P.; and Ba, J. 2014.
\newblock Adam: A method for stochastic optimization.
\newblock \emph{arXiv preprint arXiv:1412.6980}.

\bibitem[{Li, Li, and Qi(2018)}]{li-etal-2018-self}
Li, C.; Li, L.; and Qi, J. 2018.
\newblock A Self-Attentive Model with Gate Mechanism for Spoken Language
  Understanding.
\newblock In \emph{Proceedings of the 2018 Conference on Empirical Methods in
  Natural Language Processing}, 3824--3833. Brussels, Belgium: Association for
  Computational Linguistics.

\bibitem[{Libovick{\'y} and Helcl(2017)}]{libovicky-helcl-2017-attention}
Libovick{\'y}, J.; and Helcl, J. 2017.
\newblock Attention Strategies for Multi-Source Sequence-to-Sequence Learning.
\newblock In \emph{Proceedings of the 55th Annual Meeting of the Association
  for Computational Linguistics (Volume 2: Short Papers)}, 196--202. Vancouver,
  Canada: Association for Computational Linguistics.

\bibitem[{Liu et~al.(2018)Liu, Ren, Shang, Gu, Peng, and
  Han}]{liu-etal-2018-efficient}
Liu, L.; Ren, X.; Shang, J.; Gu, X.; Peng, J.; and Han, J. 2018.
\newblock Efficient Contextualized Representation: Language Model Pruning for
  Sequence Labeling.
\newblock In \emph{Proceedings of the 2018 Conference on Empirical Methods in
  Natural Language Processing}, 1215--1225. Brussels, Belgium: Association for
  Computational Linguistics.

\bibitem[{Liu et~al.(2019)Liu, Ott, Goyal, Du, Joshi, Chen, Levy, Lewis,
  Zettlemoyer, and Stoyanov}]{liu2019roberta}
Liu, Y.; Ott, M.; Goyal, N.; Du, J.; Joshi, M.; Chen, D.; Levy, O.; Lewis, M.;
  Zettlemoyer, L.; and Stoyanov, V. 2019.
\newblock Roberta: A robustly optimized bert pretraining approach.
\newblock \emph{arXiv preprint arXiv:1907.11692}.

\bibitem[{Luong, Pham, and Manning(2015)}]{luong-etal-2015-effective}
Luong, T.; Pham, H.; and Manning, C.~D. 2015.
\newblock Effective Approaches to Attention-based Neural Machine Translation.
\newblock In \emph{Proceedings of the 2015 Conference on Empirical Methods in
  Natural Language Processing}, 1412--1421. Lisbon, Portugal: Association for
  Computational Linguistics.

\bibitem[{Nguyen and Okatani(2018)}]{nguyen2018improved}
Nguyen, D.-K.; and Okatani, T. 2018.
\newblock Improved fusion of visual and language representations by dense
  symmetric co-attention for visual question answering.
\newblock In \emph{Proceedings of the IEEE Conference on Computer Vision and
  Pattern Recognition}, 6087--6096.

\bibitem[{Qin et~al.(2019)Qin, Che, Li, Wen, and Liu}]{qin-etal-2019-stack}
Qin, L.; Che, W.; Li, Y.; Wen, H.; and Liu, T. 2019.
\newblock A Stack-Propagation Framework with Token-Level Intent Detection for
  Spoken Language Understanding.
\newblock In \emph{Proceedings of the 2019 Conference on Empirical Methods in
  Natural Language Processing and the 9th International Joint Conference on
  Natural Language Processing (EMNLP-IJCNLP)}, 2078--2087. Hong Kong, China:
  Association for Computational Linguistics.

\bibitem[{Qin et~al.(2021{\natexlab{a}})Qin, Che, Ni, Li, and
  Liu}]{qin2021knowing}
Qin, L.; Che, W.; Ni, M.; Li, Y.; and Liu, T. 2021{\natexlab{a}}.
\newblock Knowing where to leverage: Context-aware graph convolutional network
  with an adaptive fusion layer for contextual spoken language understanding.
\newblock \emph{IEEE/ACM Transactions on Audio, Speech, and Language
  Processing}, 29: 1280--1289.

\bibitem[{Qin et~al.(2021{\natexlab{b}})Qin, Liu, Che, Kang, Zhao, and
  Liu}]{qin2021co}
Qin, L.; Liu, T.; Che, W.; Kang, B.; Zhao, S.; and Liu, T. 2021{\natexlab{b}}.
\newblock A co-interactive transformer for joint slot filling and intent
  detection.
\newblock In \emph{ICASSP 2021-2021 IEEE International Conference on Acoustics,
  Speech and Signal Processing (ICASSP)}, 8193--8197. IEEE.

\bibitem[{Qin et~al.(2021{\natexlab{c}})Qin, Wei, Xie, Xu, Che, and
  Liu}]{qin-etal-2021-gl}
Qin, L.; Wei, F.; Xie, T.; Xu, X.; Che, W.; and Liu, T. 2021{\natexlab{c}}.
\newblock {GL}-{GIN}: Fast and Accurate Non-Autoregressive Model for Joint
  Multiple Intent Detection and Slot Filling.
\newblock In \emph{Proceedings of the 59th Annual Meeting of the Association
  for Computational Linguistics and the 11th International Joint Conference on
  Natural Language Processing (Volume 1: Long Papers)}, 178--188. Online:
  Association for Computational Linguistics.

\bibitem[{Qin et~al.(2021{\natexlab{d}})Qin, Xie, Che, and Liu}]{ijcai2021-622}
Qin, L.; Xie, T.; Che, W.; and Liu, T. 2021{\natexlab{d}}.
\newblock A Survey on Spoken Language Understanding: Recent Advances and New
  Frontiers.
\newblock In Zhou, Z.-H., ed., \emph{Proceedings of the Thirtieth International
  Joint Conference on Artificial Intelligence, {IJCAI-21}}, 4577--4584.
  International Joint Conferences on Artificial Intelligence Organization.
\newblock Survey Track.

\bibitem[{Qin et~al.(2020)Qin, Xu, Che, and Liu}]{qin-etal-2020-agif}
Qin, L.; Xu, X.; Che, W.; and Liu, T. 2020.
\newblock {AGIF}: An Adaptive Graph-Interactive Framework for Joint Multiple
  Intent Detection and Slot Filling.
\newblock In \emph{Findings of the Association for Computational Linguistics:
  EMNLP 2020}, 1807--1816. Online: Association for Computational Linguistics.

\bibitem[{Song et~al.(2020)Song, Wang, Zhang, Zhao, Liu, and
  Liu}]{song-etal-2020-profile}
Song, H.; Wang, Y.; Zhang, W.-N.; Zhao, Z.; Liu, T.; and Liu, X. 2020.
\newblock Profile Consistency Identification for Open-domain Dialogue Agents.
\newblock In \emph{Proceedings of the 2020 Conference on Empirical Methods in
  Natural Language Processing (EMNLP)}, 6651--6662. Online: Association for
  Computational Linguistics.

\bibitem[{Srinivasan et~al.(2020)Srinivasan, Sanabria, Metze, and
  Elliott}]{Srinivasan2020MultimodalSR}
Srinivasan, T.; Sanabria, R.; Metze, F.; and Elliott, D. 2020.
\newblock Multimodal Speech Recognition with Unstructured Audio Masking.
\newblock \emph{ArXiv}, abs/2010.08642.

\bibitem[{Su, Yuan, and Chen(2018)}]{su-etal-2018-time}
Su, S.-Y.; Yuan, P.-C.; and Chen, Y.-N. 2018.
\newblock How Time Matters: Learning Time-Decay Attention for Contextual Spoken
  Language Understanding in Dialogues.
\newblock In \emph{Proceedings of the 2018 Conference of the North {A}merican
  Chapter of the Association for Computational Linguistics: Human Language
  Technologies, Volume 1 (Long Papers)}, 2133--2142. New Orleans, Louisiana:
  Association for Computational Linguistics.

\bibitem[{Teng et~al.(2021)Teng, Qin, Che, Zhao, and Liu}]{Teng2021InjectingWI}
Teng, D.; Qin, L.; Che, W.; Zhao, S.; and Liu, T. 2021.
\newblock Injecting Word Information with Multi-Level Word Adapter for Chinese
  Spoken Language Understanding.
\newblock \emph{ICASSP 2021 - 2021 IEEE International Conference on Acoustics,
  Speech and Signal Processing (ICASSP)}, 8188--8192.

\bibitem[{Tsai et~al.(2019)Tsai, Bai, Liang, Kolter, Morency, and
  Salakhutdinov}]{tsai-etal-2019-multimodal}
Tsai, Y.-H.~H.; Bai, S.; Liang, P.~P.; Kolter, J.~Z.; Morency, L.-P.; and
  Salakhutdinov, R. 2019.
\newblock Multimodal Transformer for Unaligned Multimodal Language Sequences.
\newblock In \emph{Proceedings of the 57th Annual Meeting of the Association
  for Computational Linguistics}, 6558--6569. Florence, Italy: Association for
  Computational Linguistics.

\bibitem[{Tur and De~Mori(2011)}]{tur2011spoken}
Tur, G.; and De~Mori, R. 2011.
\newblock \emph{Spoken language understanding: Systems for extracting semantic
  information from speech}.
\newblock John Wiley \& Sons.

\bibitem[{Wang, Shen, and Jin(2018)}]{wang-etal-2018-bi}
Wang, Y.; Shen, Y.; and Jin, H. 2018.
\newblock A Bi-Model Based {RNN} Semantic Frame Parsing Model for Intent
  Detection and Slot Filling.
\newblock In \emph{Proceedings of the 2018 Conference of the North {A}merican
  Chapter of the Association for Computational Linguistics: Human Language
  Technologies, Volume 2 (Short Papers)}, 309--314. New Orleans, Louisiana:
  Association for Computational Linguistics.

\bibitem[{Wu et~al.(2018)Wu, Dai, Yin, Huang, and Chen}]{wu2018improving}
Wu, Z.; Dai, X.-Y.; Yin, C.; Huang, S.; and Chen, J. 2018.
\newblock Improving review representations with user attention and product
  attention for sentiment classification.
\newblock In \emph{Thirty-second AAAI conference on artificial intelligence}.

\bibitem[{Xu and Sarikaya(2014)}]{xu2014contextual}
Xu, P.; and Sarikaya, R. 2014.
\newblock Contextual domain classification in spoken language understanding
  systems using recurrent neural network.
\newblock In \emph{2014 IEEE International Conference on Acoustics, Speech and
  Signal Processing (ICASSP)}, 136--140. IEEE.

\bibitem[{Yang et~al.(2019)Yang, Dai, Yang, Carbonell, Salakhutdinov, and
  Le}]{yang2019xlnet}
Yang, Z.; Dai, Z.; Yang, Y.; Carbonell, J.; Salakhutdinov, R.~R.; and Le, Q.~V.
  2019.
\newblock Xlnet: Generalized autoregressive pretraining for language
  understanding.
\newblock In \emph{Advances in neural information processing systems},
  5753--5763.

\bibitem[{Young et~al.(2013)Young, Ga{\v{s}}i{\'c}, Thomson, and
  Williams}]{young2013pomdp}
Young, S.; Ga{\v{s}}i{\'c}, M.; Thomson, B.; and Williams, J.~D. 2013.
\newblock Pomdp-based statistical spoken dialog systems: A review.
\newblock \emph{Proceedings of the IEEE}, 101(5): 1160--1179.

\bibitem[{Zadeh et~al.(2017)Zadeh, Chen, Poria, Cambria, and
  Morency}]{zadeh-etal-2017-tensor}
Zadeh, A.; Chen, M.; Poria, S.; Cambria, E.; and Morency, L.-P. 2017.
\newblock Tensor Fusion Network for Multimodal Sentiment Analysis.
\newblock In \emph{Proceedings of the 2017 Conference on Empirical Methods in
  Natural Language Processing}, 1103--1114. Copenhagen, Denmark: Association
  for Computational Linguistics.

\bibitem[{Zhang et~al.(2018)Zhang, Dinan, Urbanek, Szlam, Kiela, and
  Weston}]{zhang-etal-2018-personalizing}
Zhang, S.; Dinan, E.; Urbanek, J.; Szlam, A.; Kiela, D.; and Weston, J. 2018.
\newblock Personalizing Dialogue Agents: {I} have a dog, do you have pets too?
\newblock In \emph{Proceedings of the 56th Annual Meeting of the Association
  for Computational Linguistics (Volume 1: Long Papers)}, 2204--2213.
  Melbourne, Australia: Association for Computational Linguistics.

\bibitem[{Zhang and Wang(2016)}]{zhang2016joint}
Zhang, X.; and Wang, H. 2016.
\newblock A joint model of intent determination and slot filling for spoken
  language understanding.
\newblock In \emph{Proceedings of the Twenty-Fifth International Joint
  Conference on Artificial Intelligence}, 2993--2999.

\bibitem[{Zheng et~al.(2019)Zheng, Chen, Huang, Liu, and
  Zhu}]{zheng2019personalized}
Zheng, Y.; Chen, G.; Huang, M.; Liu, S.; and Zhu, X. 2019.
\newblock Personalized dialogue generation with diversified traits.
\newblock \emph{arXiv preprint arXiv:1901.09672}.

\bibitem[{Zhong, Xiong, and Socher(2018)}]{zhong-etal-2018-global}
Zhong, V.; Xiong, C.; and Socher, R. 2018.
\newblock Global-Locally Self-Attentive Encoder for Dialogue State Tracking.
\newblock In \emph{Proceedings of the 56th Annual Meeting of the Association
  for Computational Linguistics (Volume 1: Long Papers)}, 1458--1467.
  Melbourne, Australia: Association for Computational Linguistics.

\bibitem[{Zhu, Cao, and Yu(2020)}]{zhu2020dual}
Zhu, S.; Cao, R.; and Yu, K. 2020.
\newblock Dual learning for semi-supervised natural language understanding.
\newblock \emph{IEEE/ACM Transactions on Audio, Speech, and Language
  Processing}, 28: 1936--1947.

\end{thebibliography}
\end{document}